\documentclass{ifacconf}
\usepackage{amsmath}
\usepackage{amssymb} 
\usepackage{bm}
\usepackage{float}
\usepackage{natbib}        % required for bibliography
\usepackage{graphics} % for pdf, bitmapped graphics files
\usepackage{graphicx}      % include this line if your document contains figures

\usepackage{float}
\usepackage{tikz}
\usepackage{booktabs}
\usepackage{comment}
\usepackage{url}
\usepackage{etoolbox}

\begin{document}
\begin{frontmatter}
\title{
A neural signed configuration distance function for path planning of picking manipulators\thanksref{footnoteinfo}
}
\thanks[footnoteinfo]{This research was supported by the \emph{Wallenberg AI, Autonomous Systems and Software Program (WASP)} funded by Knut and Alice Wallenberg Foundation.}

\author[First]{Bernhard~Wullt} 
\author[Second]{Per~Mattsson} 
\author[Second]{Thomas~B.~Schön}
\author[First]{Mikael~Norrlöf}
\address[First]{ABB robotics, Västerås, Sweden, e-mail: name.surname@se.abb.com.}
\address[Second]{Department of Information Technology, Uppsala University, Uppsala, Sweden, e-mail: name.surname@it.uu.se}

\begin{abstract}                % Abstract of 50--100 words
Picking manipulators are task specific robots, with fewer degrees of freedom compared to general-purpose manipulators, and are heavily used in industry. The efficiency of the picking robots is highly dependent on the path planning solution, which is commonly based on sampling-based multi-query methods. The planner is robustly able to solve the problem, but its heavy use of collision-detection limits the planning capabilities for online use. We approach this problem by presenting a novel implicit obstacle representation for path planning, a neural signed configuration distance function (nSCDF), which allows us to form collision-free balls in the configuration space. We use the ball representation to re-formulate a state of the art multi-query path planner, i.e., instead of points, we use balls in the graph. Our planner returns a collision-free corridor, which allows us to use convex programming to produce optimized paths. From our numerical experiments, we observe that our planner produces paths that are close to those from an asymptotically optimal path planner, in significantly less time.
\end{abstract}
\end{frontmatter}

\newcommand{\norm}[1]{\lVert#1\rVert}

% vectors, matricies
\def\q{\textbf{q}}
\def\qc{\textbf{q}_{\text{c}}}
\def\qb{\textbf{q}_{\text{b}}}
\def\qd{\dot{\q}}
\def\qseg{\tilde{\q}}

\def\ql{\q^{\text{l}}}
\def\qu{\q^{\text{u}}}

\def\qdl{\dot{\q}^{\text{l}}}
\def\qdu{\dot{\q}^{\text{u}}}

\def\qdd{\ddot{\q}}
\def\qfree{\q_{\text{free}}}

\def\nnw{\bm{\theta}}
\def\a{\textbf{a}}
\def\b{\textbf{b}}
\def\Bmat{\textbf{B}}
\def\x{\textbf{x}}
\def\D{\textbf{D}}

\def\u{\textbf{u}}
\def\l{\textbf{l}}

\def\c{\textbf{c}}
\def\cobs{\c^{\text{obs}}}
\def\k{\textbf{k}}
\def\z{\textbf{z}}
\def\w{\textbf{w}}
\def\p{\textbf{p}}
\def\Pmat{\textbf{P}}
\def\d{\textbf{d}}
\def\r{\textbf{r}}

\def\Rmat{\textbf{R}}

\def\A{\textbf{A}}
\def\Hess{\textbf{H}}
\def\Q{\textbf{Q}}

\def\Th{\bm{\Theta}}
\def\ThOne{\bm{\Theta}_1}
\def\ThTwo{\bm{\Theta}_2}
\def\ThThree{\bm{\Theta}_3}
\def\ThThreeSrt{\bm{\tilde{\Theta}}_3}

\def\Rotz{\textbf{R}_z}

% special states
\def\xStart{\x_{\text{s}}}
\def\xGoal{\x_{\text{g}}}

\def\qStart{\q_{\text{s}}}
\def\qdStart{\dot{\q}_{\text{s}}}
\def\qGoal{\q_{\text{g}}}
\def\qdGoal{\dot{\q}_{\text{g}}}

% sets, spaces etc.
\def\C{\mathcal{C}}
\def\Cf{\C_{\text{f}}}
\def\Co{\C_{\text{o}}}
\def\Cob{\partial \Co}

\def\setFK{\mathcal{FK}}
\def\setB{\mathcal{B}}
\def\S{\mathcal{B}}

\def\W{\mathcal{W}}
\def\DS{\mathcal{D}}

\def\Z{\mathcal{Z}}
\def\Zf{\Z_{\text{free}}}
\def\Zo{\Z_{\text{obst}}}

\def\Uspace{\mathcal{U}}
\def\Xspace{\mathcal{X}}
\def\R{\mathbb{R}}
\def\tube{\tau}

\def\O{\mathcal{O}}
\def\Osta{\mathcal{O}^{\text{s}}}
\def\Odyn{\mathcal{O}^{\text{d}}}

\def\DS{\mathcal{D}}

\def\Graph{\mathcal{G}}
\def\V{\mathcal{V}}
\def\E{\mathcal{E}}

% tuples, sequences
\def\vert{\text{v}}
\def\obst{\text{o}}
\def\s{\text{s}}
\def\centers{\text{C}}

%functions
\def\funcSD{\psi}
\def\funcD{\phi}
\def\funcMSE{\text{MSE}}

% names
\def\Ast{A$^\star$}

%scalars, constants

\def\bzNrC{n}
\def\bzD{d}

\def\NrDOF{D}
\def\NrThThree{N_{u}}
\def\NrObst{N_{m}}
\def\NrBz{N_{k}}
\def\NrDisc{N_{e}}
\def\NrSampledDataPoints{N_{f}}
\def\NrSampledObstacles{N_{o}}
\def\NrVerts{N_{\V}}
\def\NrVertsCover{\tilde{N}_{\V}}
\def\NrSamples{N_{s}}

\def\rthr{\tilde{r}}
\def\timeTraj{T_{\q}}

\def\thf{\theta_2^\text{f}}
\def\thc{\theta_2^\text{c}}
\def\thb{\theta_2^\text{b}}

% distributions
\def\uniform{\mathcal{U}}
\def\geom{\textbf{g}}
\def\NrVertsExtra{N_{\V+}}
\def\NrNeighConnect{N_{\text{nn}}}

\def\Vrw{\tilde{\V}}
\def\Erw{\tilde{\E}}
\def\GraphRw{\tilde{\Graph}}

\def\geomClass{\mathcal{G}}
\def\funcNSD{\funcSD_{\textbf{w}}}

\maketitle

\newbool{arxiv}
\setbool{arxiv}{true}
\newif\ifappendix
\appendixtrue

\newcommand{\BW}[1]{{\color{red} #1}}

\section{Introduction}
Material handling such as pick and place is a common robot application in industry, with high importance for supply chain efficiency. Due to its importance, specialized manipulators like palletizing or SCARA robots have been designed for the task. These robots typically have 3 or 4 degrees of freedom (DOF), which is less than the 6 DOF for a standard industrial robot, but it is enough for the task. In a high-performing picking solution, just having a stream-lined manipulator is not enough. Of equal importance is the path planner, which should return collision-free paths at short latency. Path planning can be non-trivial for picking robots, since they can be deployed in cluttered environments or integrated tightly together as a multi-robotic system.
\\\\
Although difficult, it can be successfully solved using sampling-based planners. These planners rely on possibly the lowest level of obstacle representation available, which is collision-free points produced by a collision-detector. Clearly, relying only on points is limiting; one might therefore wonder if there are not richer representations which would alleviate some of the limitations and computational bottlenecks in the planning. Expressing new representations in the configuration space is not an easy task, since the mapping of world-space obstacles to configuration space obstacle regions is non-trivial to express analytically.
\\\\
In this paper, we present a simple approach to produce richer implicit obstacle representations in the configuration space, which we then use to re-formulate a practical multi-query path planner for picking application, resulting in a path planner which produces short paths fast. The planner is available for download\footnote{\url{https://github.com/berwul/nSCDF_PBRM}}. Specifically, our contributions are as follows: 
\begin{itemize}
    \item A simple approach to learn a signed distance function expressed in the configuration space. The learned distance function enables us to represent collision-free regions as balls. It supports an arbitrary number of obstacles and does not have to be re-trained if the work space changes. Furthermore, the distance function is differentiable, allowing us to increase the ball volume by gradient search.
    \item A multi-query path planner where we cover the free space with collision-free balls instead of points. This representation yields a collision-free corridor, which can then be used in conjunction with convex programming to, e.g., generate smooth trajectories that maximize clearance or minimize time duration.
    \item A hybrid solution, where our distance function is integrated together with a conventional collision-detector, allowing our planner to be used for higher dimensional picking problems.
\end{itemize}
%\vspace{4pt}
\begin{figure*}
\centering
\begin{tikzpicture}
\scalebox{.7}{\node[inner sep=0pt] (cs) at (0, 0) {\includegraphics[height=.33\textwidth, trim={0cm 0cm 0cm 0cm},clip]{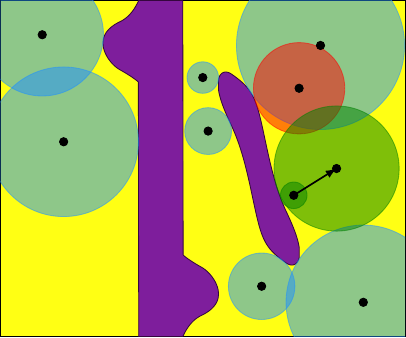}};}
\scalebox{.7}{\node[inner sep=0pt] (ws) at (8, 0) {\includegraphics[height=.33\textwidth, trim={0cm 0cm 0cm 0cm},clip]{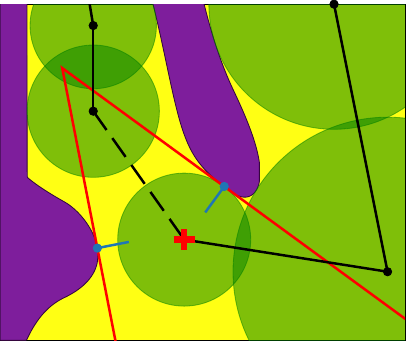}};}
\scalebox{.7}{\node[inner sep=0pt] (ws) at (16, 0) {\includegraphics[height=.33\textwidth, trim={0cm 0cm 0cm 0cm},clip]{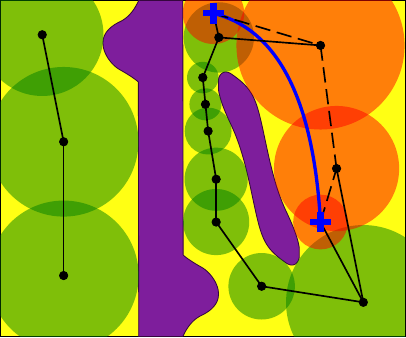}};}
\end{tikzpicture}
\vspace{-40pt}
\caption{\textbf{All}: Yellow and purple regions represents $\Cf$ and $\Co$, respectively. \textbf{Left:} Illustrates how the balls are distributed. Balls are added sequentially. Already distributed balls are colored blue. A sampled ball center within an already covered region is rejected (red ball). Once the balls have been distributed, their volumes are increased by stepping along the gradient (green balls and black arrow). \textbf{Middle:} Zoomed-in image which shows how we form the edges. First, edges between all intersecting balls are formed (solid black lines). To form global edges, we compute an approximate collision-free polytope (red lines) around each ball center (red cross). The polytope is formed by computing the closest boundary points (blue points) from the nSCDF, along with the corresponding gradients (blue lines). For each neighbouring ball that is not already connected and has a center point within the polytope, we check if its center point can be connected with a collision-free line segment using the nSCDF. If so, an edge is formed between them (dashed black line). \textbf{Right:} When we solve for online queries, we first connect the query points (blue crosses) to the graph. Then, we find a shortest path through the center points of the graph (dashed black lines). Next, we compute a sequence of overlapping balls (colored red) that make up a collision-free corridor. Within this region, we can use convex programming to compute an optimal path (blue curve) w.r.t. path length, smoothness, clearance, etc.
}
\label{fig:pbrm}
\end{figure*}
\newpage
\section{Related work}
\textit{Path planning for manipulators} can be approached in many ways \citep{robModPlanCont}. Grid-based methods discretize the free-space into a grid from which paths can be extracted \citep{Astar}, however, the approach scales bad with increasing DOFs. Optimization based methods \citep{chomp, schulman, itomp, stomp} solve an optimization problem, where distances to obstacles are expressed through a signed distance function (SDF) expressed in the world space. However, since the distance gets mapped to the configuration space by the non-linear forward kinematics, the optimization problem turns into a non-linear program, which is computationally expensive to solve and depends on a good initial guess. Possibly, the most successful approach for manipulator path planning is to use sampling-based methods \citep{sbpAnnualReview}.
\\\\
\textit{Sampling based path planners} can be dived into single-query and multi-query planners. Rapidly-exploring random trees (RRT) \citep{rrt}, is a single query planner, thus, it searches for a path from scratch each time it is queried. Probabilistic roadmap (PRM) is a multi-query planner, it starts by growing a roadmap as an initial offline step, then it solves for multiple queries by re-using the roadmap. For picking scenarios, solving multiple queries is common, which motivates using PRM as a planning approach. PRM is a \textit{probabilistically} complete planner, meaning that it is guaranteed to return a path, or inform that a path does not exist, with unit probability as the number of samples goes to infinity. To address the lack of optimality guarantees from PRM \citet{rrtStar} introduced its \textit{asymptotically} optimal counterpart PRM$^\star$. The returned paths approach the optimal paths as the number of samples in the roadmap approaches infinity. Although it comes with optimality guarantees, the main drawback is that the approach is computationally expensive, therefore making it unsuitable for picking scenarios where quality paths need to be returned fast. Motivated by the great practical benefits of PRM, we enhance it by using collision-free regions in the form of balls in the configuration space, which has the main benefit of enabling convex optimization.
\\\\
\textit{Collision-free configuration space regions} in motion planning is covered in \citep{sbng, mpGcs}. Similar to our approach \citet{sbng} proposes a graph covered with configuration-space balls. The use case is, however, different, where they focus on feedback motion. Furthermore, the balls are formed from upper bounds on link displacements and world-space distances, making the approach conservative. Since gradient information is missing, it also becomes hard to increase the ball volume further. \citet{mpGcs} relies instead on polytopes to describe the free-space, paths are then produced by using the graph of convex sets (GCS) \citep{gcs} framework, which finds the shortest path in the graph by solving a relaxed mixed-integer convex program. To produce polytopes, results from \citet{IRISnlp, irisC} is used, which utilizes optimization methods to iteratively grow a polytope in the configuration space. Compiling a graph can then be done with, e.g, \citet{vcc}. Although an attractive approach, computing polytopes through optimization is computationally heavy, which naturally affects the graph compilation step, making the overall offline computation time too heavy to be practically useful. Recent work addresses this by relying on GPUs \citep{werner2025superfast}, which, however, is not always available for a robot system. Furthermore, since a relaxed problem is solved in GCS, it may produce invalid paths, making it necessary to solve a mixed-integer convex program, which is too computationally heavy to solve online. In our approach, we also utilize convex optimization, but instead of trying to find a path globally in the graph, we decompose the problem. First, we find a rough path through the center points in the balls. We then refine it by opening up a convex corridor, where we are free to optimize for, e.g., smoothness \citep{convexSmoothing}, trajectory duration \citep{biconvex} or whatever our application requires. This decomposition, i.e., finding a rough path which is then refined, is a common strategy in path planning \citep{lavallebook}. A recent work similar to ours is presented by \citet{fastPathPlanning}, which also uses convex programming to optimize a trajectory. However, the planning is only done in world space. Finally, concurrent work to ours is given in \citep{lee2024safe}, where sampling-based planning using distance functions is proposed. Planning is, again, only conducted in world-space. We go beyond this, focusing on planning in configuration space, which is where the actual planning takes place for manipulators.
\\\\
\textit{Learning based collision detection} is done  in \citep{fkKernelDistance, diffco, graphdistnet}, where the idea is to predict the signed distance between the robot links and the surrounding obstacles in the world space. This speeds up computations compared to conventional methods \citep{GJK}. Our approach is instead to learn an SDF expressed in the configuration space, which allows us to form the largest collision-free ball region around a given configuration. A similar learning approach is presented in \citep{cspf}. Their approach differs by learning only a distance function, therefore requiring a conventional collision-detector at run-time. We learn a \textit{signed} distance function, thus collision status is included in the learned function. Secondly, their data process is computationally more expensive, two functions need to be learned to generate the data. Our approach directly finds configurations on the boundary by collision-detection. Thirdly, their approach is restricted to point cloud representations, which becomes computationally heavy at deployment. We propose to represent the obstacles as parameterized geometric shapes, e.g. spheres or axis-aligned boxes. This reduces computation time, since a large number of points can potentially be represented by a single obstacle. Finally, we demonstrate how to utilize distances to express collision-free regions and propose a novel path planner based on this concept.
\section{Problem formulation}
A robot is located in the world space, $\W \subset \R^3 $. The unique location of the robot is given by its configuration $\q \in \C \subset \R^D$, where $\C$ is the configuration space. We focus on picking applications, thus $D\le4$. The set of points covered by the robot's bodies at a certain configuration is denoted as $\setFK(\q) \subset \W$. The robot is surrounded by $n$ obstacles $\O = \bigcup_{i=1}^{n} \O_i \subset \W$. The representation of the obstacle in the configuration space is the set $\C\O_i = \{\q \in \C \: |\: \setFK(\q) \cap \O_i \neq \emptyset \}$. The obstacle space is formed as $\Co = \bigcup_{i=1}^n \C \O_i$ and by the complement  $\Cf = \C \setminus \Co$ we get the free space. The path planning problem is to connect a query pair, i.e. a start, $\qStart$, and goal configuration, $\qGoal$, with a geometric path, $\q(s): [0, 1] \mapsto \Cf$, such that $\q(0)=\qStart$ and $\q(1)=\qGoal$, or report correctly when such a path does not exist.
\section{Method}
\subsection{Signed configuration distances}
\begin{figure*}
\centering
\begin{tikzpicture}
\scalebox{.8}{\node[inner sep=0pt] (ws) at (0, 0) {\includegraphics[height=.33\textwidth, trim={11cm 1cm 5cm 4cm},clip]{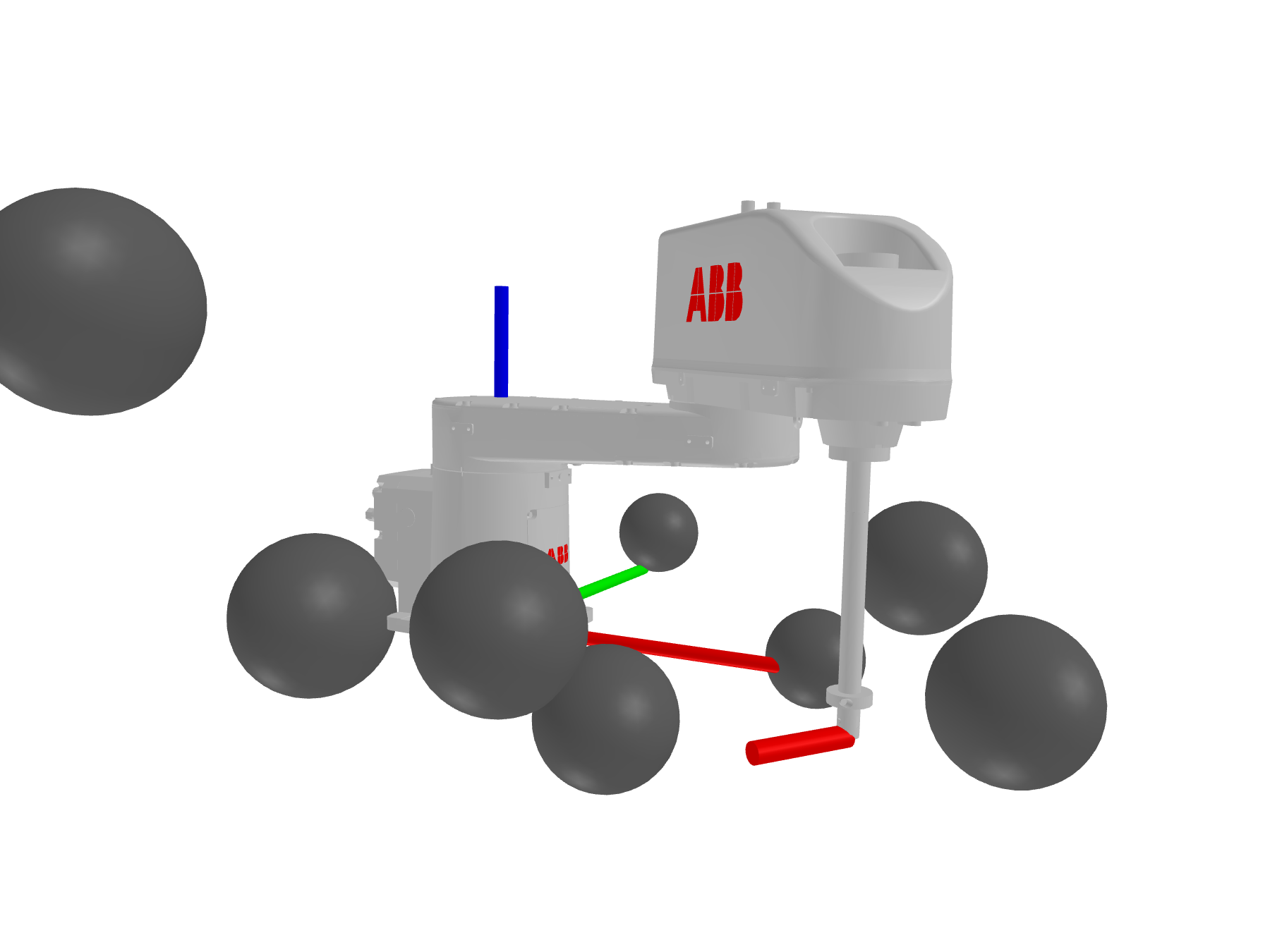}};}
\scalebox{.8}{\node[inner sep=0pt] (cs) at (7,0) {\includegraphics[height=.33\textwidth, trim={5cm 0 5cm 0},clip]{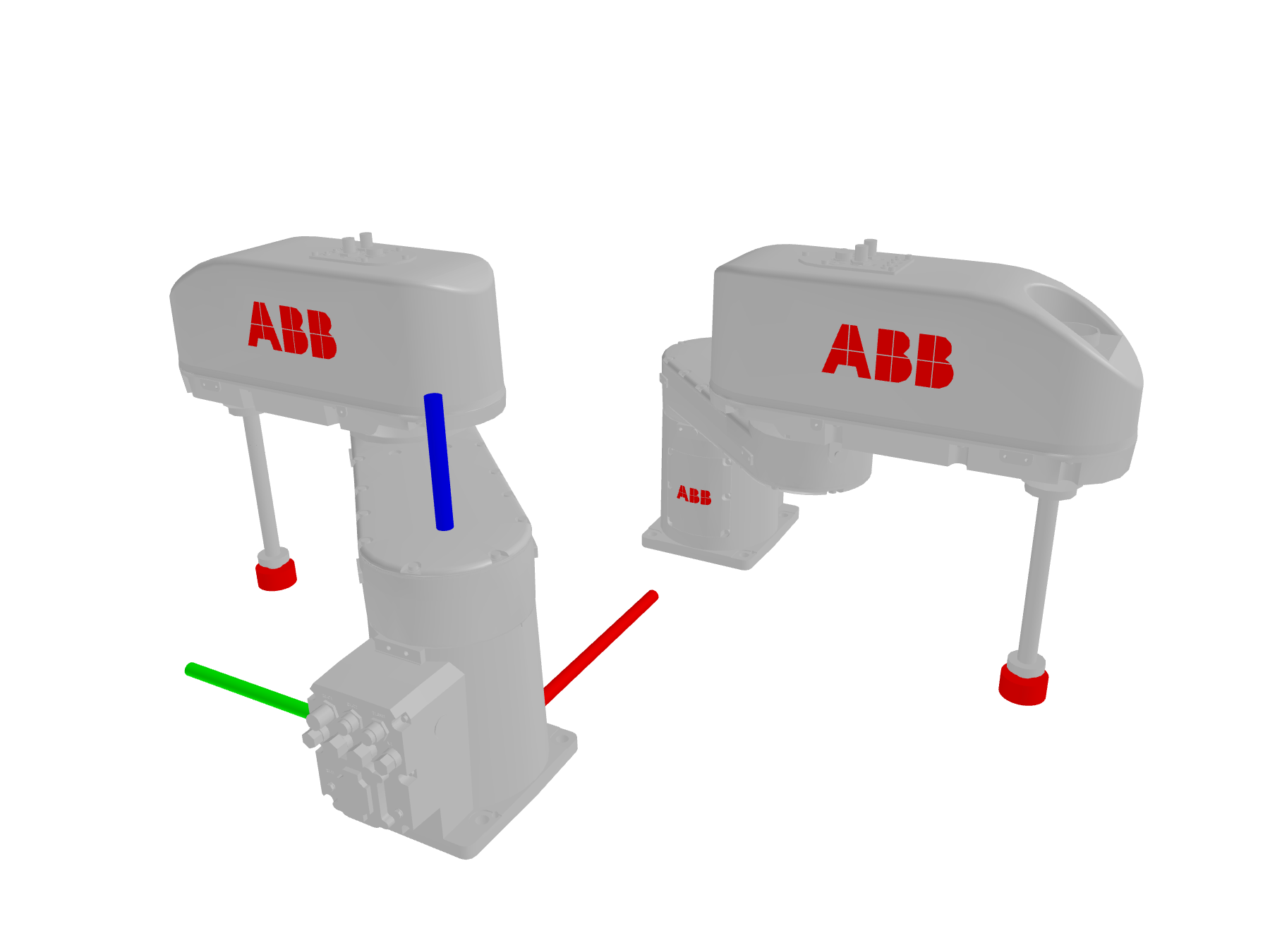}};}
\scalebox{.8}{\node[inner sep=0pt] (ws) at (14, 0) {\includegraphics[height=.33\textwidth, trim={10cm 0 5cm 0},clip]{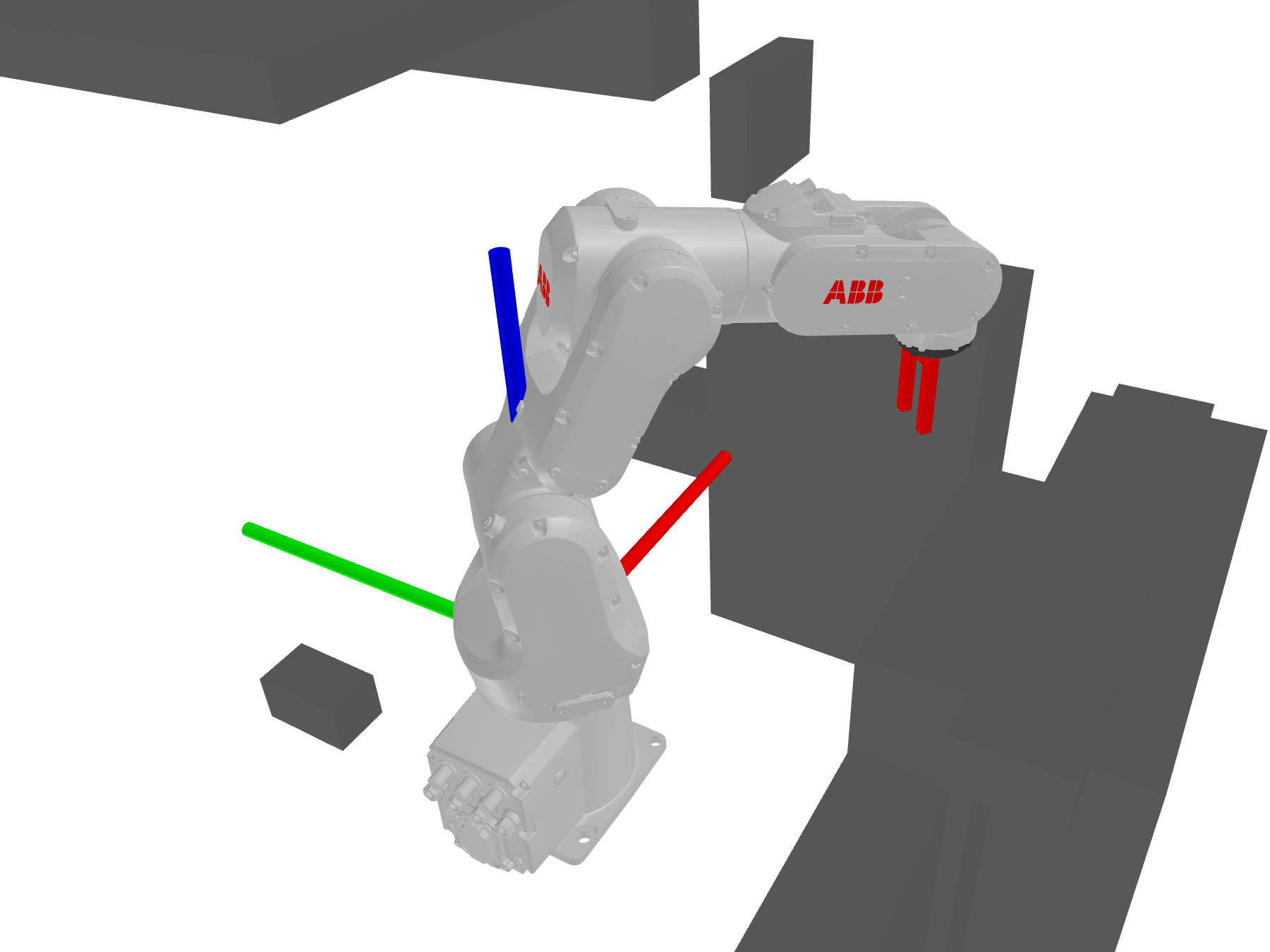}};}

\node[text width=1cm] at (0.25, -2) {\textbf{SC3/SC4}};
\node[text width=1cm] at (5.5, -2) {\textbf{MR}};
\node[text width=1cm] at (10.75, -2) {\textbf{WD}};
\end{tikzpicture}
\vspace{-30pt}
\caption{\textbf{All:} The coordinate systems are colored red, green, and blue for $x$, $y$ and $z$ axes, respectively. Tools are colored red. Obstacles are colored gray. For all systems a self-collision nSCDF is learned in addition to the obstacle representation. \textbf{SC3/SC4:} SCARA robot with spherical obstacles (with and without tool). \textbf{MR:} Multi-robot system with two SCARAs. \textbf{WD:} 6 DOF robot kinematically constrained in wrist down configuration, resulting in 3 DOF. Obstacles are axis-aligned boxes.}
\label{fig:experiments_systems}
\end{figure*}
\label{sec:NSCDF}
\subsubsection{Background} We define the configuration distance function, $\funcD: \C \times \W \mapsto \R_+ $, as
\begin{equation}
\funcD(\q, \O_i) = \min_{\qc \in \partial\C\O_i} \: \norm{\q - \qc},
\end{equation}
where $\norm{\cdot}\mapsto \R_+$ is the 2-norm. It takes a configuration, $\q$, and a world-space representation of the obstacle, $\O_i$, as input and returns the smallest distance to the boundary of the obstacle region in the configuration space, $\partial\C\O_i$. The distance function gives us no information on whether we are in collision or not. This information is instead embedded in the \textit{signed} configuration distance function (SCDF), $\funcSD: \C \times \W \mapsto: \R $, defined as 
\begin{equation}
\funcSD(\q, \: \O_i) = 
\begin{cases}
-\funcD(\q, \: \O_i) & \text{if } \q \in \C\O_i, \\
\hphantom{-} \funcD(\q, \: \O_i) & \text{otherwise.}
\end{cases}
\end{equation}
Thus, if $\q$ is in the interior of $\C\O_i$, i.e., in collision, then the distance is negative; otherwise it is positive, i.e., collision-free. The SCDF allows us to represent collision-free regions in the form of balls, that is
\begin{equation}
    \label{eq:ball}
    \setB(\c) = \{ \q \in \R^\NrDOF  \: | \:   \norm{\q - \c} \le \funcSD(\c) \} \subset \Cf.
\end{equation}
Furthermore, if the SCDF is a smooth function, then we can grow the ball volume, i.e., the collision-free region, by stepping along the gradient, $\nabla_{\c} \funcSD(\c, \O_i)$, since it points in the direction of increasing distance.
\\\\
From the problem description, we realize that our problem scenario involves multiple obstacles, $\O=\bigcup_{i=1}^{n} \O_i$, which in general can result in a more complicated geometry. A naive approach would be to form an SCDF for the resulting shape, which seems to be a harder problem than forming an SCDF for the individual shapes. Fortunately, the SCDF allows us to combine multiple SCDF values, from possibly multiple different SCDFs, by taking the smallest distance, that is
\begin{equation}
\label{eq:scdf:min}
\funcSD(\q, \O) =  \text{min}(\funcSD_1(\q, \O_1), \: \hdots, \: \funcSD_m(\q, \O_{n})).
\end{equation}
Thus, the SCDF inherently supports multiple obstacles, which makes it flexible and scalable. The attractive properties makes it a good tool for path planning. However, the non-linear mapping of a world-space obstacle to its corresponding obstacle boundary in the configuration space is far from clear how to form analytically.
\subsubsection{Neural implicit representation} 
For our application, it is possible to numerically compute an approximate SCDF, see\ifbool{arxiv}{~Appendix~\ref{app:numerical_SCDF}}{~\cite{arxiv_version_of_paper}}. However, the process is computationally expensive; therefore, to reduce the computation time, our proposed approach is to approximate the SCDF with a neural network, $\funcSD \approx \funcNSD$, which we refer to as a neural SCDF (nSCDF). Since we are using a neural network to approximate the SCDF, we need to be able to represent a general obstacle, $\O_i$, in parametric form so that it can be fed into the network. Therefore, we restrict the nSCDF to support geometrical shapes that come from the same family of geometries, e.g., the family of spheres, parameterized by a vector $\geom \in \geomClass \subset \R^G $, where $\geomClass$ is the parameter space. Thus, the nSCDF is defined as the mapping $\funcNSD(\q, \: \geom) : \C \times \geomClass \mapsto \R $. Arbitrary obstacles can be represented as a union of possibly simpler shapes, e.g., union of spheres \citep{hubbardSpheres}, and the resulting SCDF is then computed by using the min-property \eqref{eq:scdf:min}. Furthermore, the flexibility which the min-property brings us, makes it possible to learn distinct different nSCDFs for different geometrical shapes, e.g., spheres, axis-aligned boxes, self-collisions, and combine them with \eqref{eq:scdf:min}. We train the neural network in a supervised learning way, learning the distance from a generated data set consisting of a large set of random obstacles and configurations, computing the signed distance according to\ifbool{arxiv}{~Appendix~\ref{app:numerical_SCDF}}{~\cite{arxiv_version_of_paper}}.
\subsection{Multi-query path planning}
In the following section, we describe how we integrate convex collision-free regions into the roadmap planner PRM. The primary benefit of this new representation is that it enables us to utilize convex programming to generate paths. There are other benefits to this representation, which we highlight in the sections below. Our approach to achieve this is simple; instead of representing collision-free points in the roadmap, we use the collision-free balls~\eqref{eq:ball} produced by the nSCDF. We refer to our representation as Probabilistic Bubble RoadMap (PBRM). The following sections starts by presenting the offline stages, i.e. vertex and edge creation, ending with the online stage.
\subsubsection{Vertices distribution}
The vertex creation process is illustrated in the left part of Figure \ref{fig:pbrm}. We start by sampling $m$ collision-free points and add them to the graph sequentially. Points that are already covered by any ball in the graph are resampled. In this way a larger portion of the free-space is covered. Once we have distributed the $m$ balls, we use the gradient information available from the nSCDF to increase the overall coverage, by growing each balls volume along the gradient direction.
\subsubsection{Edge creation}
The edge computation is illustrated in the middle part of Figure \ref{fig:pbrm}. Compared to PRM, our representation gives us connectivity information for free; we simply form edges between any pair of balls that intersect. To further increase the connectivity beyond the local scope of each vertex, we compute an approximate convex collision-free region around its center point. This is achieved by computing an approximate separating hyperplane for each obstacle, which is formed from the gradient and the configuration distance associated with that obstacle. Using all hyperplanes from all obstacles results in a polytopic region that is approximately collision-free. We use this region to find neighboring points that have not already been connected by collision-checking all transitions with the nSCDF, adding an edge if the transition is collision-free. The benefit of this approach, compared to the more standard approach like connecting within a fixed radius, is that we eliminate an additional tuning parameter and directly use the geometrical information available from the nSCDF when estimating our neighbors.
\subsubsection{Query phase}
Once the graph is compiled, we can solve for arbitrary queries, i.e., start and goal pairs, $\qStart$ and $\qGoal$, presented in the right part of Figure \ref{fig:pbrm}. The first step is to connect the points to the graph. For each point, we compute its ball representation \eqref{eq:ball} and check for intersections with the balls in the roadmap. If there are intersections, we add the edges to the graph; otherwise, we use the standard approach in PRM, where we try to connect $m$ neighbours by a collision-free straight line using the nSCDF. After connecting the query points to the graph, we solve for the shortest path in the graph by running A$^\star$, resulting in a sequence of overlapping balls $(\setB_1, \hdots, \setB_{k})$, which represents a collision-free corridor that connects the query points. The corridor enables us to use convex optimization to, e.g., compute a smooth trajectory or optimize for path length.
\section{Experiments}
\label{sec:exp}
\subsubsection{Multi-query path planning}
\begin{table}[t]
\centering
\caption{The computation time and path lengths are mean values. Our planner results are divided into the initial path along ball centers (PBRM) and the optimized path in the corridor (PBRM$^\star$).}
\label{tab:static:results}
\begin{tabular}{l|cc|cc}
 & \multicolumn{2}{c}{\textbf{MR}} & \multicolumn{2}{c}{\textbf{SC3}} \\
\midrule
 & Time [ms] &  Length [-] & Time [ms] &  Length [-] \\
\midrule
PBRM &  \textbf{0.62} & 2.37 &  \textbf{1.85} & 2.20 \\
PBRM$^\star$ & 2.77 &  \textbf{1.71}  &  4.59 & 1.78  \\
PRM & 115.58 & 2.04 & 111.72 & 1.86\\
PRM$^\star$ & 1014.36 & 1.84 & 1013.29 &  \textbf{1.66} \\
\midrule
 & \multicolumn{2}{c}{\textbf{SC4}} & \multicolumn{2}{c}{\textbf{WD}} \\
\midrule
PBRM &  \textbf{3.92} & 2.94 &  \textbf{0.95} & 1.31  \\
PBRM$^\star$ &7.84 & 2.49   & 3.19 &  \textbf{1.19} \\
PRM & 119.74 & 2.63  &  122.86 & 1.36  \\
PRM$^\star$ & 1015.21 &  \textbf{2.24} & 1019.46 & 1.24 \\
\end{tabular}
\end{table}
The focus of this section is to present the benefits of the proposed implicit obstacle representation in path planning. To show the wide range of obstacle representations it can handle, we learn an nSCDF for four different scenarios, illustrated in Figure~\ref{fig:experiments_systems}, which also includes self-collisions. The scenarios are more thoroughly described in\ifbool{arxiv}{~Appendix~\ref{app:systems}. The details of the learning and validation process are presented in~Appendix~\ref{app:learning}}{~\cite{arxiv_version_of_paper} together with the learning and validation process}.
\\\\
Having learned the nSCDFs for the different systems, we next benchmark our proposed planner with the state-of-the-art multi-query planners PRM and PRM$^\star$ using the implementation from \citet{ompl}. The benchmark is done through randomized experiments with details explained in\ifbool{arxiv}{~Appendix \ref{app:pp_setup}}{~\cite{arxiv_version_of_paper}}. It is emphasized that no learning occurs at this stage and the obstacles used have not been encountered during training. We measure performance based on the path length and the time it takes to produce a path. 
\\\\
The results are presented in Table \ref{tab:static:results}. Clearly, PBRM has the lowest compute time, however, the paths are by far the longest. PRBM$^\star$ optimizes a path in the corridor, solving the convex program presented in\ifbool{arxiv}{~Appendix~\ref{app:convex_opt}}{~\cite{arxiv_version_of_paper}}, resulting in significantly reduced path lengths at a relatively low computational cost, reaching path lengths close to PRM$^\star$, even shorter in some cases. The success rate, i.e., the fraction of instances when a path was returned, was overall high, in the range of 90-100\%, and more or less the same between the planners. An exception for this was observed in the 4 DOF SCARA scenario, where our planner was roughly 10 percentage points lower in success rate compared to the benchmarks. The reason for this was that the randomized world instances resulted in narrow passages, and as observed in\ifbool{arxiv}{~Appendix \ref{app:learning}}{~\cite{arxiv_version_of_paper}}, our learned nSCDF is conservative, therefore making it difficult to connect the graph. Regarding collisions, the collision-rate was low and more or less the same for all planners.
\subsubsection{Hybrid path planning}
\begin{figure}
\centering
\includegraphics[height=.3\textwidth, trim={8cm 0cm 0cm 0cm},clip]{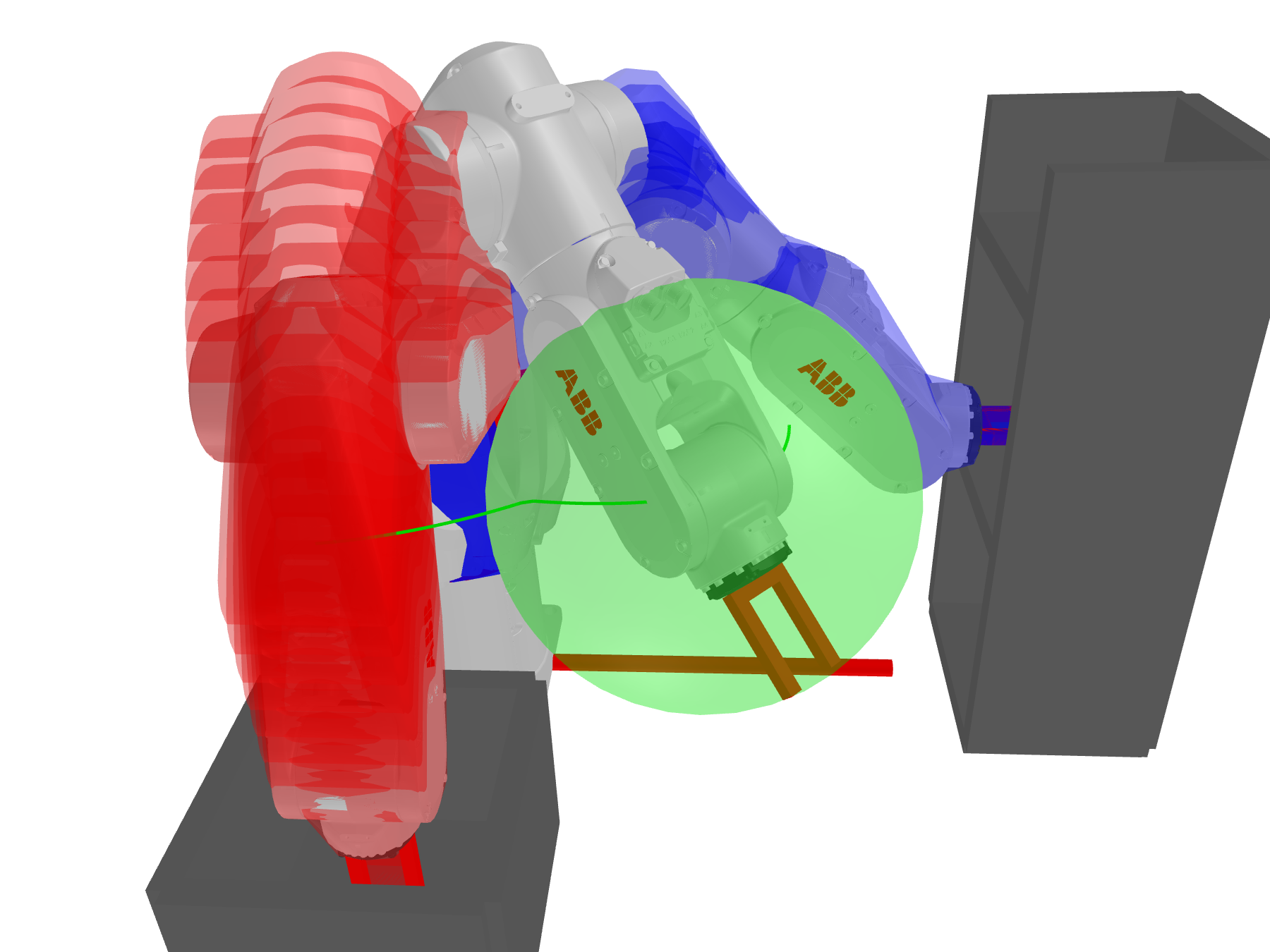}
\caption{A demonstration of how to use the PBRM in a hybrid solution for a typical pick and place scenario.}
\label{fig:demo}
\end{figure}
Finally, we demonstrate how our planner can be used together with a conventional collision-detector in a hybrid way, making it capable of solving higher dimensional planning problems. We showcase this by solving the scenario presented in Figure \ref{fig:demo}. It illustrates a picking scenario, where we pick an object from a box and place it on a shelf. We solve it by first learning an nSCDF for axis-aligned boxes, where we replace the tool with a bounding sphere (transparent green sphere). If a configuration has a positive nSCDF distance, this means that we are allowed to freely select a configuration for the wrist axes, i.e., the last three joints. We initialize the planner as previously described. To connect a query point with negative nSCDF to a point in the graph, we fall back on a conventional collision-detector. If it is collision-free, then we increment along the edge, otherwise we stop and try to connect with another point in the graph. We form an edge between the points once a complete edge has been verified to be collision-free. The figure illustrates how the start and goal configurations are connected through this process, where the transparent red and blue meshes illustrate where a collision-detector was used. The average time to connect a query point to the graph for the showcased scenario was 2 ms. Once a path is found, we can open up a collision-free corridor over the path segments where the nSCDF is positive and produce an optimized path. The green curve in the figure illustrates the traced-out position of the wrist center point along the optimized path. The total time to produce the path was 10 ms, which, compared to the times reported in Table~\ref{tab:static:results}, is still relatively fast, despite the extra computation time from the conventional collision-detector.
\\\\
Naturally, this approach becomes unsuitable for scenarios that are very cluttered. However, for scenarios where this is not the case, our outlined approach can be used to boost the planning.
\section{Conclusions and future work}
Path planning for picking applications is a challenging problem, where current planners are limited in their capabilities to produce paths. We identified that better representations are available in the form of signed distances; however, they are too computationally expensive to produce online. Our approach was therefore to train a neural network in a supervised way to speed up the online process. With the collision-free regions from the network, we proposed a novel multi-query path planner, that has the main benefit of returning a collision-free corridor consisting of convex balls, allowing us to use convex optimization to swiftly produce an optimized path. We evaluated our learned distance function on four different scenarios, observing good generalization capabilities. Then we evaluated our path planner in randomized experiments. Our planner returned paths in the shortest time with a wide margin. Further optimizing the path in the corridor resulted in path lengths that were near the asymptotically optimal path planner PRM$^\star$, but computed at the fraction of the time. Finally, we also showed how our planner could be used to solve higher dimensional planning problems, by integrating the nSCDF with a conventional collision-detector, resulting in a hybrid solution. Overall, our suggested planner offers a good balance between path length and computation time, which is beneficial for picking applications, where high-quality paths are expected to be returned fast. Future work will investigate how multiple nSCDFs from motion primitives can be combined for path planning.
\bibliography{ifacconf}
\ifappendix
\newpage
\appendix
\section{Numerical approximation of SCDF}
\label{app:numerical_SCDF}
We achieve a numerical approximation by discretizing the configuration space into a 4-dimensional grid, with $N$ points in each dimension, $\bm{\theta} = (\theta_1, \hdots, \theta_{N})$. We compute the collision status for all the points in the grid. Then we loop over all 3-dimensional sub-grids, which each have a fixed value of the fourth joint. From all points in collision we compute a mesh, $\mathcal{M}_i \subset \R^3$. If there are no configurations in collision, we simply continue iterating. At the end of the process, we end up with $M \le N$ meshes, $(\mathcal{M}_1, \hdots, \mathcal{M}_M)$, displaced along the axis of the 4-th joint with corresponding displacements $(\theta_1, \hdots, \theta_M)$.
\\\\
To compute a signed configuration distance for an arbitrary configuration $\q$, we compute the smallest squared distance to all the meshes positioned in $\C$, which is
\begin{equation}
    d^2 = \min_{i \in (1, \hdots, M) }  \: ( ( q_4 - \theta_i)^2  \: + \: \min_{\c \in \mathcal{M}_i } || \q_{1:3}  -  \c ||_2^2 ).
\end{equation}
In the above, $q_4$, denotes selecting the fourth element from $\q$, while $\q_{1:3}$ denotes selecting elements 1 to 3. Having computed the smallest distance, we then compute the collision status $ m \in \{ -1, 1 \}$ and the signed distance is formed as $m \cdot d$.
\section{Systems}
\label{app:systems}
\subsubsection{SCARA and spherical obstacles}
The SCARA robot has 4 DOF. If the tool is rotationally symmetric w.r.t. its last joint, then the configuration space effectively reduces to 3 DOF. We learn an nSCDF for both cases. The obstacles are spherical, thus the geometry vector is described as $\geom=(\p, \: r) \in \R^4 $, where $\p \in \R^3$ is the center position and $r \in \R_+$ its radius.
\subsubsection{Multi-robot system}
The geometry vector is described as $\geom=\begin{bmatrix} p \end{bmatrix}$, where $p \in \R_+$ is the displacement along the $x$-axis between the base links of the robots. A naive approach would consider all DOFs of the combined system, which is 8 DOF. However, in this scenario, we assume that the tool or object being picked is small. Since the robots are aligned in the $xy$-plane, collision can therefore only occur with the main links, reducing the configuration space to 4 DOFs.
\subsubsection{IRB 1100 wrist down with AAB}
The robot is kinematically constrained in a wrist-down configuration, which is commonly used in pick and place, assembly, machine tending, etc. Concretely, this means that the wrist axis always points down, which has the effect of reducing the path planning problem to 3 DOF. The geometry vector is described as $\geom=(\b, \: \d) \in \R^6$, where $\b \in \R^3$ is the center position of the box and $\d \in \R^3_+$ the box dimensions.
\subsubsection{Self-collisions}
For all systems, we also compute signed distances for self-collisions. The process is the same as for obstacles. The geometry vector is defined as the empty vector $\geom=(\cdot) \in \emptyset $.
\section{Learning details}
\label{app:learning}
\subsubsection{Data generation}
For all systems, except the multi-robot system, we generated 4096 random obstacles in the work space of the robot and computed 2048 signed distances uniformly within the robot's joint limits. We use a grid size of 32 for each dimension. For the multi-robot system, 32 different horizontal displacements were sampled, for each displacement, roughly $3 \cdot 10^5$ signed distances were computed. Thus, for each system, a data set of roughly $8\cdot 10^6$ data points was used. Generating the data for each system took roughly 2 hours on a standard laptop with an Intel i5-1155G7 processor (2.50 GHz).
\subsubsection{Network architecture} We use a basic neural network architecture consisting of four linear layers. We use $128$ and $32$ hidden units when learning distances for obstacles and self-collision, respectively. ReLU is used as an activation function for all hidden layers. In the last output layer, no activation is used. We use a mean squared error as loss function. The data set is split with a $90/10\%$ ratio. The training is stopped after 30 consecutive epochs with no improvement in test loss. The network with the lowest test loss is selected for further evaluation. The networks were trained on a system equipped with an NVIDIA TITAN Xp GPU (12 GB VRAM). The average training time was roughly 2 hours.
\subsubsection{Validation}
To validate the learned nSCDF, we sample 10 random geometry vectors that were not seen during the learning. For each geometry vector, we grid the configuration space and compute the collision status. Then, we query the nSCDF for the signed distances, and define a configuration as being in collision if the distance is negative. Performance is then evaluated by computing the accuracy and recall. The results for the different systems are presented in Table~\ref{tab:inference}.
\begin{table}[H]
\centering
\caption{Statistics from validation obstacles.}
\label{tab:inference}
\begin{tabular}{lcccc}
System            & \textbf{SC3}   & \textbf{SC4} & \textbf{MR}     & \textbf{WD} \\
\midrule
Accuracy [\%]     & 97                & 98              & 100                & 98 \\
Recall [\%]       & 98                & 94              & 99                 & 91 \\
Precision [\%]    & 68                & 80              & 87                 & 76 
\end{tabular}
\end{table}
Observing the results, we see that the overall accuracy is high. The recall (accuracy among the configurations in collision) is higher than $90 \%$ for all scenarios, which is deemed to be good enough for our purposes. The precision (accuracy over the predicted collisions) fluctuates somewhat between the systems, but is more or less higher than $70 \%$. The lower precision indicates that the nSCDF is conservative. 
\section{Experiments setup}
\label{app:pp_setup}
\subsubsection{World space instances}
For all systems, we sample 10 random obstacles, i.e., geometry vectors. For each world instance, we sample 10 random collision-free queries. Furthermore, to avoid trivial queries, we require that the query pairs cannot be connected with a collision-free line. 
\subsubsection{Benchmarks}
We benchmark our planner against the original PRM and its asymptotically optimal version PRM$^\star$, using the implementation from OMPL~\citep{ompl}. In the offline construction of the roadmaps, we allow our planner to distribute 250 vertices, while the benchmarking planners are given a competitive advantage of distributing 500 vertices. We note that the extra number of vertices for the benchmarks do not contribute to a significant increase in computation time during the online phase. The dominating computation step is to verify the edges through collision-detection when the query points are connected to their nearest neighbours. Thus, we limit all planners to check for a maximum of 10 nearest neighbors in the online phase. For PRM$^\star$, we limit the online search time to $1\:s$ to get good convergence. PBRM uses the nSCDF for collision detection. We validate all paths with a conventional collision detector. Experiments were performed on a computer with an Intel i5-1155G7 processor (2.50 GHz), thus no GPU was used to speed up the neural network computations.
\section{Shortest path in corridor}
\label{app:convex_opt}
The PBRM returns a corridor, which allows us to run a convex program to produce an optimal path within the corridor. For the experiments, we solve the optimization problem
\begin{subequations}
\label{eq:cvx}
\begin{alignat}{2}
&\!\underset{\x_1, \hdots, \x_{k+1}}{\min} &\qquad& \sum_{i=1}^{k} \norm{\x_{i} - \x_{i+1}} \label{eq:cvx:obj} \\
&\text{subject to} 	&& \x_1 = \qStart, \: \x_{k+1} = \qGoal, \label{eq:cvx:query} \\
&&& \x_{i} \in \setB_{i-1} \cap \setB_{i}, \: i = 2, \hdots k. \label{eq:cvx:int}
\end{alignat}
\end{subequations}
The objective in the above optimization problem \eqref{eq:cvx:obj} is to produce the shortest path, that starts in the query pairs \eqref{eq:cvx:query} and with intermediate points constrained to lie in the intersections between the balls \eqref{eq:cvx:int}. The optimization problem is a quadratically constrained quadratic program, which we solve with CVXPY~\citep{cvxpy}.
\fi
\end{document}